\documentclass{article}
\PassOptionsToPackage{numbers,sort&compress}{natbib}

\usepackage[preprint]{neurips_2026}
\usepackage[utf8]{inputenc}
\usepackage[T1]{fontenc}

\usepackage[hidelinks,hypertexnames=false]{hyperref}
\usepackage{url}
\usepackage{booktabs}
\usepackage{graphicx}
\usepackage{float}
\usepackage{amsfonts}
\usepackage{amsmath}
\usepackage{microtype}
\usepackage{xcolor}
\newcommand{\lit}[1]{\textcolor{black!48}{#1}}
\setcitestyle{square,comma}
\newcommand{\projecturl}{https://github.com/realAllenSong/Learning-When-to-Think-While-Listening}

\title{Learning When to Think While Listening in Large Audio-Language Models}
\author{%
  Zhiyuan Song \quad Weici Zhao \quad Yang Xiao \quad Suhao Yu \quad Cheng Zhu \quad
  Jiatao Gu\thanks{Corresponding author.}\\
  University of Pennsylvania
}

\begin{document}

\maketitle

\begin{abstract}
Recent advances in Large Audio-Language Models (LALMs) have made real-time,
streaming spoken interaction increasingly practical. In this setting, reasoning
quality and responsiveness are tightly coupled: delaying reasoning until the
speech endpoint can improve answer quality but moves deliberation into
user-visible response delay, while answering too early risks committing before
decisive evidence arrives. We introduce a learnable wait-think-answer control
formulation for LALMs. Motivated by the incremental nature of human
conversation, the controller decides under partial audio evidence when to wait,
when to externalize a compact reasoning update, and when to answer. Using Qwen2.5-Omni-7B as the base model, we construct aligned wait-think-answer traces
from spoken reasoning data, train the controller with supervised fine-tuning
(SFT), and then apply Decoupled Clip and Dynamic Sampling Policy Optimization
(DAPO). The reward combines answer
correctness, action validity, update timing, latency synchronization, reasoning
quality, and chain consistency, optimizing the complete wait-think-answer
trajectory and not the final answer alone. On a six-task synthetic spoken reasoning question answering (SRQA) benchmark, the six-reward DAPO controller improves the row-weighted accuracy from 67.6\% to 70.3\% while reducing post-endpoint final-think length by 14\% under the same Qwen deployment harness. On a 186-item human-recorded Real Audio Bench, a transfer check beyond text-to-speech (TTS)-rendered speech, the controller family remains functional: SFT achieves the strongest accuracy, while the six-reward DAPO controller is the only learned variant whose final-think length falls below the base. These results suggest that a streaming model should learn
when to make intermediate reasoning explicit during the audio stream. The
public repository is available on \href{\projecturl}{GitHub}.

\end{abstract}

\section{Introduction}
\label{sec:intro}

Large Audio-Language Models (LALMs) are moving audio research from offline
perception tasks toward real-time spoken interaction. Recent systems can
process streaming audio and generate text or speech with low delay
\cite{moshi2024,miniomni2024,freezeomni2024,qwen2025omni,qwen3omni2025},
making it increasingly realistic to ask an audio model to reason during live
speech, before the utterance has finished. In this setting, answer
quality and responsiveness are coupled: a spoken assistant that answers
correctly only after a long silent delay can still fail as an interactive
system, while a fast response can be wrong if late-arriving evidence changes the
answer.

Reasoning-intensive speech exposes a basic control problem. Most spoken
reasoning evaluations follow a complete-audio protocol: the model receives the
full spoken question, generates a reasoning trace, and then produces a response.
This protocol is useful for measuring final answer quality, but it removes the
online decision faced by a streaming LALM. At each time step in the audio stream,
the model has partial acoustic and linguistic evidence plus whatever reasoning
state it has already emitted. Waiting until the speech endpoint can preserve
accuracy, but it moves deliberation into the user-visible response delay;
answering early is responsive, but risks committing before decisive evidence
arrives. The central question is when the model should update its reasoning,
alongside what the reasoning should contain.

This temporal question is natural in human conversation, where listeners often
prepare responses before the current turn has ended and where short response
gaps require overlap between comprehension and production planning
\cite{sacks1974turntaking,stivers2009turntaking,levinson2015turntaking,bogels2015neural,castellucci2022speechplanning,stephens2010coupling}.
It is also becoming explicit in speech-model research. Chain-of-thought and
reasoning-oriented post-training improve complex problem solving in text and
audio models
\cite{wei2022cot,deepseekr12025,dapo2025,audiocot2025,audioreasoner2025,audiothinker2025},
while recent streaming systems ask whether models can think while listening or
interleave reasoning with speech generation
\cite{shih2026speech,stitch2026,shanks2025,streamingthinker2025}. We build on
this direction but train a different primitive: a stateful wait-think-answer
controller whose visible thoughts accumulate across streaming audio windows,
instead of a single post-audio rationale or a question-completeness trigger.

We introduce a learnable wait-think-answer formulation for streaming speech
reasoning. In this formulation, the LALM is an online controller whose
serialized action space contains three primitives:
\texttt{<wait/>}, \texttt{<think>...</think>}, and
\texttt{<answer>...</answer>}. A \texttt{wait} action consumes more audio
without changing the visible reasoning state. A \texttt{think} action emits a
grounded intermediate update, and an \texttt{answer} action commits to the final
response. Restricting the action space to these three primitives keeps timing
trainable and inspectable while leaving the model free to generate open-ended
text inside thinking and answering spans. Distributing reasoning across
pre-endpoint think actions amortizes the total reasoning cost: each visible
update emitted during listening absorbs deliberation that would otherwise
accumulate into the final post-endpoint think, reducing residual
post-endpoint deliberation as a controller-level proxy for user-visible
response delay.

We instantiate this controller on Qwen2.5-Omni-7B \cite{qwen2025omni}. The data
pipeline constructs aligned wait-think-answer traces from spoken reasoning
examples, renders the user input as speech, aligns text actions to audio time,
and converts each example into chained streaming windows. We first train the
controller with supervised fine-tuning (SFT), then use the resulting controller
to initialize Decoupled Clip and Dynamic Sampling Policy Optimization (DAPO).
The reward combines answer correctness, action validity, update timing, latency
synchronization, thought quality, and chain consistency, so optimization covers
the complete streaming trajectory.

The experiments focus on the resulting accuracy--residual-latency trade-off
under the deployment protocol. On synthetic spoken reasoning, the six-reward
DAPO controller improves row-weighted accuracy from 67.6\% for the base
deployment controller to 70.3\%, while reducing post-endpoint final-think
length from 10.44 to 8.99 tokens. We also evaluate the controllers on Real
Audio Bench, a 186-item human-recorded transfer check. There, SFT gives the
strongest accuracy, while DAPO variants expose shorter-reasoning operating
points. The six-reward variant is the only learned controller whose final-think
length falls below the base deployment controller. These results support the
main claim that online LALM reasoning can be trained as wait-think-answer
control: the model learns when to wait and when to externalize intermediate
reasoning during listening, while answer commitment is endpoint-gated in this
paper.

The paper contributes a training formulation for wait-think-answer control in
LALMs, focusing on the learning problem and controller semantics, not
optimized cache-native serving: a learned policy over partial audio evidence,
an aligned spoken-controller data pipeline with supervised fine-tuning and
policy optimization, and a reward that scores correctness, timing, latency,
and thought consistency. We
evaluate the resulting controllers on synthetic spoken reasoning and
human-recorded audio, showing how post-training moves the controller along
the accuracy--residual-latency frontier.

\begin{figure}[t]
\centering
\includegraphics[width=\linewidth]{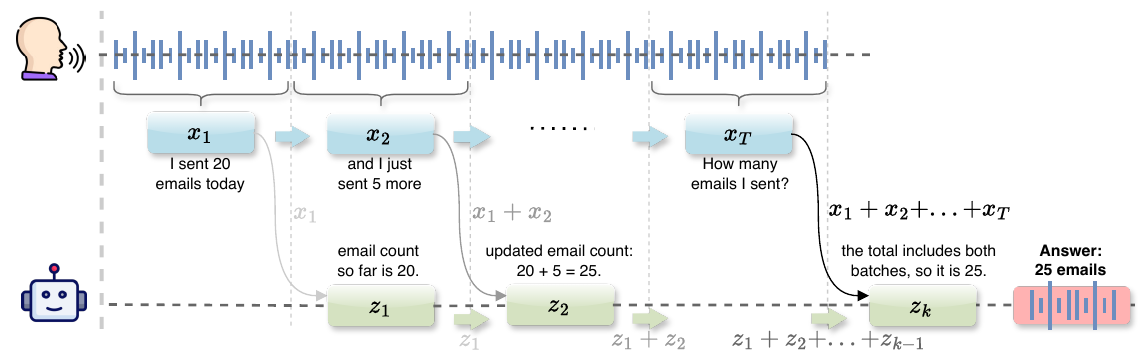}
\caption{Wait-think-answer controller in full-prefix mode. At decision step
$k$, the controller observes the audio prefix $x_{1:t_k}$ and the committed
text memory $z_{<k}$, then emits \texttt{<wait/>}, a visible state update, or a
final answer. The diagram focuses on controller-level information flow, not the internal multimodal token
layout.}
\label{fig:controller_overview}
\end{figure}

\section{Related work}
\label{sec:related}

We review three lines of work: large audio-language and omni models, audio
chain-of-thought reasoning, and online spoken interaction.

\subsection{Large audio-language and omni models}
LALMs extend text large language models (LLMs)
with audio encoders and cross-modal alignment modules, allowing speech, sound,
and music to be used as model context. Surveys of this area describe a rapid
shift from task-specific audio systems toward general instruction-following
models for audio
understanding and dialogue \cite{almsurvey2025,holisticevalsurvey2025}.
Qwen-Audio and Qwen2-Audio are representative large-scale audio-language
systems trained across heterogeneous speech, sound, and music tasks
\cite{qwenaudio2023,qwen2audio2024}. More recent omni models aim to remove the
cascaded automatic speech recognition (ASR)--LLM--text-to-speech (TTS)
boundary. Moshi uses parallel speech streams and an inner monologue channel
\cite{moshi2024}, Mini-Omni and Freeze-Omni target
low-latency speech-to-speech dialogue \cite{miniomni2024,freezeomni2024}, and
Qwen2.5-Omni introduces a streaming Thinker--Talker architecture that perceives
audio, vision, and text while generating text and speech \cite{qwen2025omni}.
We use this omni-model setting as the substrate, but study a different problem:
how the model should update an explicit reasoning state while streaming audio is
still arriving.

\subsection{Audio chain-of-thought reasoning and post-training}
Chain-of-thought (CoT) prompting established that explicit intermediate reasoning can
improve text LLM performance on arithmetic, commonsense, and symbolic tasks
\cite{wei2022cot}. Audio-CoT transfers this idea to LALMs and finds that CoT can
help easy and medium audio reasoning tasks, while hard tasks can suffer when the
reasoning chain is low quality or unnecessarily long \cite{audiocot2025}.
Subsequent audio reasoning systems improve the reasoning process through
structured data, curriculum learning, or reward-based post-training
\cite{audioreasoner2025,sari2025,audioflamingosoundcot2025,audsemthinker2025,r1aqa2025,omnir12025,stepaudior12025}.
Audio-Thinker studies adaptive audio reasoning by deciding whether to produce a
final pre-answer rationale and by training the reasoning process to stay
consistent \cite{audiothinker2025}. Our problem is different because the
controller must act before the complete utterance is available. It repeatedly
decides under partial audio evidence whether to wait, emit a visible state
update, or answer, instead of choosing only whether to reason after hearing the
audio. SFT teaches this action language. DAPO,
building on recent reasoning reinforcement learning methods
\cite{deepseekr12025,dapo2025}, then optimizes the whole streaming trajectory
for answer correctness, update timing, thought quality, and response latency.

\subsection{Thinking while listening and turn-taking}
Human conversation is fast enough that listeners often cannot wait for a turn to
finish before preparing a response. Conversation analysis and cross-linguistic
studies characterize turn-taking as a locally managed system that avoids both
overlap and long silence \cite{sacks1974turntaking,stivers2009turntaking}.
Levinson and Torreira argue that such short response gaps imply overlap between
comprehension and production planning \cite{levinson2015turntaking}.
Electroencephalography (EEG) and speech-planning studies provide evidence that response preparation can begin
before the current speaker finishes
\cite{bogels2015neural,magyari2017temporal,castellucci2022speechplanning}.
Speaker--listener neural coupling further links successful communication to
shared, time-evolving representations \cite{stephens2010coupling}. Recent speech-LLM
work turns this observation into an engineering problem. Shih et al. ask whether speech LLMs can think while listening: at inference
time, a question-completeness check (using entropy decrease over the
streaming ASR distribution) decides when the user has finished speaking,
and the model then emits a single post-utterance chain of thought, with the
accuracy--latency trade-off shaped by length-biased preference optimization
\cite{shih2026speech}. STITCH, SHANKS, and StreamingThinker study related
forms of simultaneous reasoning over speech or text streams
\cite{stitch2026,shanks2025,streamingthinker2025}. The wait-think-answer
formulation follows this motivation but trains the controller end-to-end to
act under partial audio: instead of a single terminal chain-of-thought
triggered by an inference-time completeness signal, the model learns to emit
multiple grounded intermediate state updates over the audio stream, each
visible to and reusable by later controller decisions.

\section{Method}
\label{sec:method}

\subsection{Architecture}
\label{sec:problem}

We treat streaming speech reasoning as an online control problem over an
explicit text memory (Figure~\ref{fig:controller_overview}). Let $x_{1:T}$
denote the spoken input, let $0<t_1<\cdots<t_K\leq T$ be controller decision
times, and let $z_{<k}$ be the visible reasoning states already emitted by
the model. We use a
\emph{full-prefix} controller observation: at decision step $k$, the audio input
is the complete prefix $x_{1:t_k}$ heard so far, including the newest chunk and
all earlier audio, while the text input is the visible reasoning state
$z_{<k}$.
The controller observation is
\begin{equation}
\label{eq:controller}
    o_k=(x_{1:t_k}, z_{<k}), \qquad
    a_k=\arg\max_{a\in\mathcal{A}_k}\pi_\theta(a \mid o_k),
\end{equation}
where $o_k$ denotes the controller observation, $a_k$ denotes the action
taken at step $k$, $\pi_\theta$ is the controller policy with parameters
$\theta$, and $\mathcal{A}_k$ is the legal action set at that step.
Equation~\eqref{eq:controller} is the deterministic deployment rule. The
underlying policy $\pi_\theta(\cdot \mid o_k)$ is a distribution over legal
actions, and DAPO training samples from it to generate rollout groups. We
omit the fixed controller instruction from the notation because it is constant
across examples. Before the speech endpoint, the legal actions are
\textsc{wait} and \textsc{think}. After the endpoint, the controller emits one
final \textsc{think} and then an \textsc{answer}. We serialize these actions
as \texttt{<wait/>}, \texttt{<think>...</think>}, and
\texttt{<answer>...</answer>}. Thus, answer timing is not a learned
pre-endpoint stopping decision in our experiments; the \textsc{answer} action
is retained to train and validate the complete interaction contract. A
\textsc{wait} advances the stream without changing memory. A pre-endpoint
\textsc{think} appends a short semantic state update $z_k$, so every later
controller call can condition on both the complete audio prefix and the
previously committed thoughts. At the endpoint, the final \textsc{think}
receives the complete audio $x_{1:T}$ and all earlier visible thoughts, and
the answer turn receives the same complete audio plus the final-think state
before emitting $\hat{y}$.

The same information pattern would appear in a cache-native streaming
deployment. In that deployment, audio chunks would keep entering one
persistent Qwen2.5-Omni context \cite{qwen2025omni}, the audio key-value (KV)
cache would not be discarded, intermediate \textsc{think} text would be
appended back into the visible context, the final-think turn would see the
complete cached audio plus all earlier thoughts, and the answer turn would see
that same context plus the final-think state. The available Qwen2.5-Omni serving path does not expose an
official controller-style cache interface for this loop, and an efficient custom
implementation would require lower-level runtime and kernel work. We study the
same information flow through full-prefix replay: every controller call replays
the observed prefix while preserving the evidence available in the intended
deployment semantics. Appendix~\ref{app:train} and
Section~\ref{sec:limitations} discuss this approximation further.

\subsection{Training data construction}

Following the spoken reasoning data format of Shih et al.~\cite{shih2026speech},
we first generate semantic controller traces with GPT-4o \cite{gpt4o2024} and
then ground them in speech. The initial candidate corpus contains 80,000
records: 40,000 verifiable items and 40,000
open-ended items. Generation is stratified by verifiability,
difficulty, and topic.
Appendix~\ref{app:prompts} gives the prompt and field schema. Each record stores
a spoken surface form, a TTS style instruction, lexical anchors for
answer-relevant state changes, a semantic wait-think-answer trace, and the final
answer.

After validation, deduplication, and human spot checks, the aligned corpus used
for the reported experiments contains 75,723 audio-text records: 38,213
verifiable and 37,510 open-ended. The train/validation split contains 73,675 and
2,048 records, respectively. SFT uses the full aligned corpus. DAPO keeps the
verifiable training branch, yielding 37,180 scorable training records. To expose
the reward to benchmark-style reasoning formats during training, we add a small
set of training-split examples from ARC-Challenge, ARC-Easy, GSM8K, PIQA, and SocialIQA. These
rows remain below 2\% of the DAPO mix and never include held-out evaluation
examples.

We synthesize each spoken input as a single full utterance with Qwen3-TTS
\cite{qwen3tts2026}. This preserves global prosody and avoids artificial breaks
at action boundaries. Connectionist temporal classification (CTC)-style forced
alignment \cite{graves2006ctc} maps transcript words to timestamps, and
controller boundaries are snapped upward to the 0.5s decision grid. This grid
controls reconsideration frequency, not the amount of audio evidence visible at
each call. Appendix~\ref{app:train} reports the tick sweep.

\subsection{Supervised fine-tuning}

We first train the controller with supervised fine-tuning on Qwen2.5-Omni-7B
\cite{qwen2025omni} using low-rank adaptation (LoRA) \cite{lora2022} in
MS-Swift \cite{swift2024}.
The supervised objective teaches the action
serialization, the short semantic state style, and the distinction between
ordinary waits, answer-relevant pre-endpoint thoughts, final-think compression,
and the final answer. The supervised controller is trained from the base
Qwen2.5-Omni-7B model for one epoch over the aligned controller export.
Training and validation curves show decreasing loss and increasing token
accuracy, with both stabilizing during training. DAPO runs are
initialized from supervised controllers trained with this recipe. In the
reported tables, the SFT row reports this supervised controller, and the DAPO
rows report four-, five-, and six-reward policy-optimization variants.
Appendix~\ref{app:train} reports the training curves, hyperparameters, and
compute details.

\subsection{DAPO policy optimization}

After SFT, we optimize the controller with a controller-specific DAPO
policy-optimization loop \cite{dapo2025}. SFT can be handled directly in
MS-Swift, but DAPO requires live streaming rollouts, controller-action parsing,
local reward computation, dynamic resampling, and adapter updates inside the
controller runtime. This requirement motivates a custom trainer instead of the
generic MS-Swift Group Relative Policy Optimization (GRPO) path. DAPO belongs
to the GRPO family \cite{deepseekmath2024,deepseekr12025}. Each prompt is sampled into a group of
$G$ rollouts scored with the trajectory reward in Eq.~\ref{eq:reward_total}.
We compute group-relative advantages and apply token-level clipped updates to
controller, thought, and answer tokens. Token-level credit is useful because the
policy mixes actions with different failure modes: bad waits, useful short
thoughts, malformed final-think turns, and correct answers should not all
receive identical pressure. Appendix~\ref{app:train} gives the objective,
clipping settings, Kullback--Leibler (KL) regularization, and protocol-gate details.

\subsection{Reward design}
\label{sec:reward}

\begin{figure}[!b]
\centering
\includegraphics[width=0.92\linewidth]{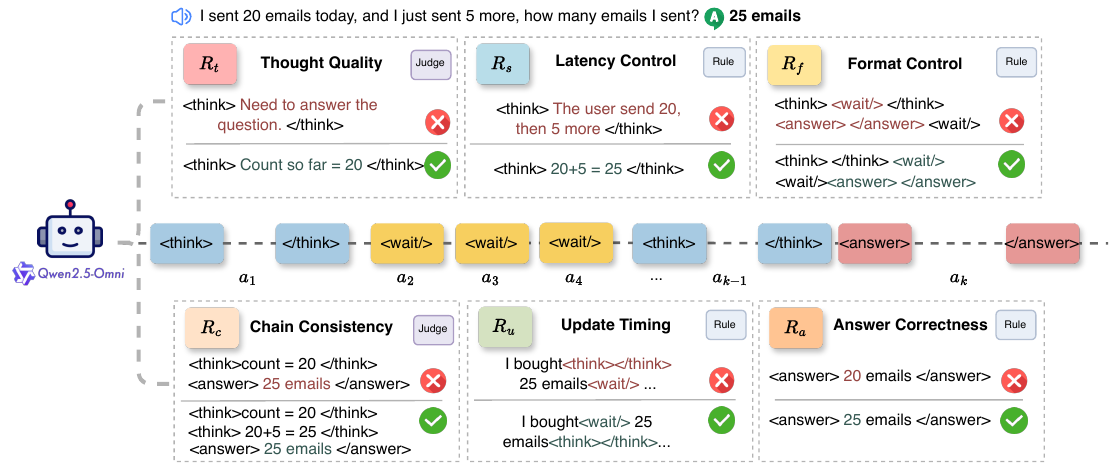}
\caption{Trajectory reward for wait-think-answer control. Rule-based terms
enforce action format, update timing, answer correctness, and final-latency
budget, while judge-assisted terms score local thought quality and chain
consistency.}
\label{fig:reward_overview}
\end{figure}

The reward targets the main failure modes of online speech reasoning.
Optimizing only the final answer encourages a wait-all policy that shifts
deliberation to the user-visible response delay. Optimizing only latency can
produce premature answers, empty thoughts, or malformed action traces. We score
the full wait-think-answer trajectory so that a good rollout needs to answer
correctly, follow the interaction format, update its state when the evidence
changes, and keep the final post-endpoint deliberation compact.
Table~\ref{tab:reward_components} summarizes the six reward terms, and
Figure~\ref{fig:reward_overview} illustrates how they score a wait-think-answer
trajectory.

\begin{table}[!t]
\caption{Reward terms for wait-think-answer control. Each term targets a
specific failure mode in the streaming trajectory, from invalid action format
to unsupported final reasoning.}
\label{tab:reward_components}
\centering
\scriptsize
\renewcommand{\arraystretch}{1.08}
\setlength{\tabcolsep}{3.0pt}
\resizebox{\linewidth}{!}{%
\begin{tabular}{@{}lll@{}}
\toprule
Term & Role & Training signal \\
\midrule
$R_a$ & Answer correctness & Rewards the final answer using deterministic task scoring whenever possible; a local judge is used only as a semantic-equivalence fallback for open short-answer cases. \\
$R_f$ & Protocol validity & Enforces the wait-think-answer contract: listen-time actions are wait/think, and the answer appears only after the completed utterance. \\
$R_s$ & Response latency & Favors compact final-think states, so reasoning already resolved during listening is not repeated after the endpoint. \\
$R_u$ & Update timing & Rewards thoughts near answer-relevant state changes and penalizes both missed updates and irrelevant interruptions. \\
$R_t$ & Thought quality & Prefers short, concrete semantic states that support the answer over generic or verbose commentary. \\
$R_c$ & Chain consistency & Checks whether the visible thought chain supports the final answer without unsupported jumps or contradictions. \\
\bottomrule
\end{tabular}
}
\end{table}

Let $\tau$ be a completed streaming trajectory. The reward combines four
rule-based terms with two judge-assisted reasoning terms. $R_a$ measures final
answer correctness through task scoring, with judge scoring used only as a
semantic-equivalence fallback for open short-answer cases. $R_f$ enforces the
controller format, $R_s$ penalizes residual final-think latency, and $R_u$
rewards well-timed state updates.
Inspired by process-level audio reasoning rewards
\cite{audiothinker2025}, $R_t$ scores whether each visible thought is short and
answer-supporting, while $R_c$ scores whether the resulting thought chain
supports the final answer. The shaped reward for a valid trajectory is
\begin{equation}
R_{\mathrm{valid}}(\tau)=
\underbrace{\lambda_a R_a}_{\text{answer}}
+\underbrace{\lambda_f R_f}_{\text{format}}
+\underbrace{\lambda_s R_s}_{\text{sync/latency}}
+\underbrace{\lambda_u R_u}_{\text{update timing}}
+\underbrace{\lambda_t R_t}_{\text{thought quality}}
+\underbrace{\mathbf{1}[R_a>0]\lambda_c R_aR_c}_{\text{consistency bonus}},
\end{equation}
and the final reward applies a protocol gate:
\begin{equation}
\label{eq:reward_total}
R(\tau)=
\begin{cases}
\lambda_f R_f, & R_f\leq 0,\\
R_{\mathrm{valid}}(\tau), & R_f>0.
\end{cases}
\end{equation}
We use
$\lambda_a=1.0,\lambda_f=1.0,\lambda_s=1.0,\lambda_u=3.0,\lambda_t=1.0$, and
$\lambda_c=0.45$. The gate gives protocol validity priority over all other
objectives: if the model answers before the utterance is complete, omits the
required final-think/answer structure, or emits an ill-formed trace, answer
correctness cannot rescue the rollout. The consistency term is also gated by
$R_a>0$, so a fluent but wrong chain does not receive extra credit simply for
being self-consistent.

The ablations in Section~\ref{sec:results} add these terms progressively. The
four rule-based terms ($R_a,R_f,R_s,R_u$) establish the basic
accuracy--residual-latency--timing trade-off. Adding $R_t$ and $R_c$ asks a stricter
question: whether improving the content of the intermediate state also improves
the ability of the controller to spend reasoning earlier in the stream and less
after the endpoint.

\section{Experiments and evaluation}
\label{sec:experiments}

\paragraph{Synthetic spoken reasoning question-answering benchmark.}
We adopt the spoken reasoning question answering (SRQA) evaluation construction introduced by Shih
et al.~\cite{shih2026speech}. Following their setup, text reasoning problems are
rewritten into spoken questions, rendered with TTS, and then evaluated as audio
inputs. Our benchmark contains 8,959 items across ARC-Easy (ARC-E),
ARC-Challenge (ARC-C) \cite{arc2018}, Physical Interaction: Question Answering (PIQA) \cite{piqa2020}, SocialIQA (SIQA)
\cite{socialiqa2019}, GSM8K \cite{gsm8k2021}, and a 300-item short-answer
factual set (LLaMA-QS) following the LLaMA-Questions/Spectron source family
\cite{spectron2023}. We follow their spoken-rewrite prompt lineage and use a
local Qwen3.6-35B-A3B judge \cite{qwen36_35b_a3b2026} for short-answer scoring,
together with task-specific normalization for numeric and multiple-choice
answers. Appendix~\ref{app:benchmark} gives task counts and conversion details.

\paragraph{Evaluation protocols.}
We report two protocols. \emph{Offline mode} is the complete-audio
setting: the model receives the full spoken question once, emits one final
\texttt{<think>...</think>} state, and then emits the answer. This protocol
measures answer quality after complete audio observation and serves as the
standard upper-bound reference for models that do not expose a controller
interface. \emph{Deployment mode} is our streaming protocol: audio arrives on a
fixed 0.5s grid, each controller call sees the full audio prefix plus previous
visible thoughts, and the policy chooses \texttt{<wait/>} or
\texttt{<think>...</think>} before the endpoint. At the endpoint, the controller emits one
final \texttt{<think>...</think>} and then an
\texttt{<answer>...</answer>}. If a controller call is still running when a
later tick arrives, stale triggers can be skipped, matching the practical
constraint that an online system cannot spend unbounded work at every trigger.

Deployment mode is only available for Qwen-family rows, because the controller
requires direct control over the Qwen2.5-Omni inference loop
\cite{qwen2025omni}: prefix replay, explicit wait/think/answer action parsing,
and persistence of intermediate thought states must all be implemented inside a
single benchmark harness. External baselines do not expose a comparable
controller interface in our setup, so we evaluate them only in offline mode. To
separate model quality from protocol effects, the base Qwen2.5-Omni-7B is
reported in both offline and deployment modes.

\paragraph{Human-recorded real audio.}
We further introduce Real Audio Bench, a human-recorded streaming benchmark
collected for this work. Five speakers recorded 200 GPT-4o-generated
\cite{gpt4o2024} candidate items designed for natural spoken delivery. Human
screening removed ambiguous or unanswerable prompts and corrected answer keys,
leaving 186 recordings. Scoring uses normalized rules together with the
same local Qwen3.6-35B-A3B judge \cite{qwen36_35b_a3b2026} for cases requiring
semantic matching. We treat this as a compact transfer benchmark for real human
delivery, not as a substitute for a large-scale user study.

\paragraph{Baselines and metrics.}
We compare Qwen-family controller stages: the base Qwen2.5-Omni-7B model
\cite{qwen2025omni}, the SFT controller, and DAPO controllers trained with
four, five, or six reward terms. The base model is reported in both offline and
deployment modes to isolate the effect of the streaming controller protocol
itself. The SFT controller provides the supervised initialization for the DAPO
policy-optimization runs. The learned controller variants are evaluated in deployment mode, which
is the setting they are trained to optimize. External baselines, including
Audio Flamingo 3 \cite{af32025}, Audio Flamingo 3 + AF-Think
\cite{audioflamingosoundcot2025}, and GLM-4-Voice-9B \cite{glm4voice2024}, are
evaluated only in offline mode because our benchmark harness cannot impose the
same controller interface on third-party models without re-implementing their
inference stacks. We also include the literature-reported Moshi rows from Shih
et al.~\cite{shih2026speech} for context.

Our main latency metric is the length of the post-endpoint final-think state,
used as the controller-level measure of residual deliberation after the user stops speaking. This
matches the interactive concern behind the latency metric of Shih et
al.~\cite{shih2026speech}, but Qwen2.5-Omni does not expose Moshi-style
time-aligned monologue tokens. We treat cross-family token-latency
comparisons as descriptive. The core latency claims compare Qwen streaming
controllers under the same harness. Because full-prefix replay also incurs
repeated prefix prefill cost, Appendix~\ref{app:runtime} reports
replay-harness real-time factor (RTF) for the full-prefix deployment audit.

\section{Results}
\label{sec:results}

\subsection{Synthetic spoken SRQA}

Table~\ref{tab:synthetic_main} reports per-task accuracy on the six-task
synthetic spoken reasoning benchmark, plus the row-weighted average and a
token-level residual-output measure. The offline rows establish the strength of
modern audio models under complete-audio evaluation. The controller rows isolate
the question optimized by wait-think-answer post-training: under the same
deployment protocol, can policy optimization improve the controller without
changing the underlying audio model?

\begin{table}[t]
\caption{Results on synthetic spoken SRQA. Task columns report accuracy (\%).
Avg. is row-weighted accuracy. Final $\downarrow$ reports the mean length of
the final post-audio \texttt{<think>...</think>} state when that span is
produced by the evaluation prompt. For grey Moshi rows, the latency metric is
copied from Shih et al.~\cite{shih2026speech} and is not directly comparable to
Qwen controller tokens. Bold and underlined values
mark the best and second-best within the Qwen streaming controllers block,
since complete-audio baselines and reported Moshi rows operate under
different evaluation protocols. QC denotes question
completeness, and DPO denotes direct preference optimization.}
\label{tab:synthetic_main}
\centering
\scriptsize
\renewcommand{\arraystretch}{0.88}
\setlength{\tabcolsep}{1.8pt}
\resizebox{\linewidth}{!}{%
\begin{tabular}{@{}lrrrrrrrr@{}}
\toprule
Method & ARC-E & ARC-C & SIQA & PIQA & GSM8K & LLaMA-QS & Avg. $\uparrow$ & Final $\downarrow$ \\
\midrule
\multicolumn{9}{@{}l}{\textit{Complete-audio baselines}} \\
Qwen2.5-Omni-7B & 89.6 & 81.1 & 70.9 & 71.3 & 26.7 & 71.0 & 70.8 & 10.27 \\
Audio Flamingo 3 & 75.5 & 59.4 & 42.2 & 32.7 & 10.1 & 66.3 & 47.4 & 4.97 \\
Audio Flamingo 3 + AF-Think & 47.3 & 35.0 & 31.6 & 21.7 & 3.0 & 48.0 & 30.5 & 9.18 \\
GLM-4-Voice-9B & 64.9 & 48.5 & 41.2 & 22.1 & 6.7 & 63.7 & 40.2 & 20.88 \\
\midrule
\multicolumn{9}{@{}l}{\lit{\textit{Reported streaming baselines from Shih et al.}}} \\
\lit{Moshi baseline} & \lit{30.2} & \lit{21.5} & \lit{22.8} & \lit{23.8} & \lit{8.7} & \lit{42.8} & \lit{23.4} & \lit{--} \\
\lit{Moshi + CoT} & \lit{77.7} & \lit{59.8} & \lit{56.1} & \lit{56.9} & \lit{16.1} & \lit{57.8} & \lit{56.6} & \lit{--} \\
\lit{Moshi + CoT w/o streaming ASR} & \lit{55.8} & \lit{44.0} & \lit{50.1} & \lit{46.3} & \lit{12.2} & \lit{59.9} & \lit{44.8} & \lit{--} \\
\lit{Moshi QC-SFT} & \lit{62.8} & \lit{43.2} & \lit{45.1} & \lit{40.7} & \lit{13.8} & \lit{56.2} & \lit{44.4} & \lit{52.42} \\
\lit{Moshi QC + length-DPO} & \lit{65.4} & \lit{46.0} & \lit{45.3} & \lit{46.0} & \lit{14.7} & \lit{56.9} & \lit{46.7} & \lit{19.31} \\
\midrule
\multicolumn{9}{@{}l}{\textit{Qwen streaming controllers}} \\
Qwen2.5-Omni-7B & 87.8 & \underline{80.8} & 68.6 & 63.5 & 22.8 & \underline{71.0} & 67.6 & 10.44 \\
SFT controller & 86.3 & 78.1 & 68.6 & 60.5 & 21.6 & \textbf{71.7} & 66.1 & \underline{9.82} \\
DAPO controller (4 rewards) & 88.9 & \textbf{81.7} & 68.4 & 65.1 & 24.6 & 70.3 & 68.5 & 10.87 \\
DAPO controller (5 rewards) & \underline{89.1} & \textbf{81.7} & \underline{69.6} & \underline{66.4} & \underline{24.9} & \underline{71.0} & \underline{69.2} & 10.94 \\
DAPO controller (6 rewards) & \textbf{89.6} & \textbf{81.7} & \textbf{71.0} & \textbf{69.2} & \textbf{25.9} & \underline{71.0} & \textbf{70.3} & \textbf{8.99} \\
\bottomrule
\end{tabular}
}
\end{table}

\begin{figure}[t]
\centering
\includegraphics[width=0.88\linewidth]{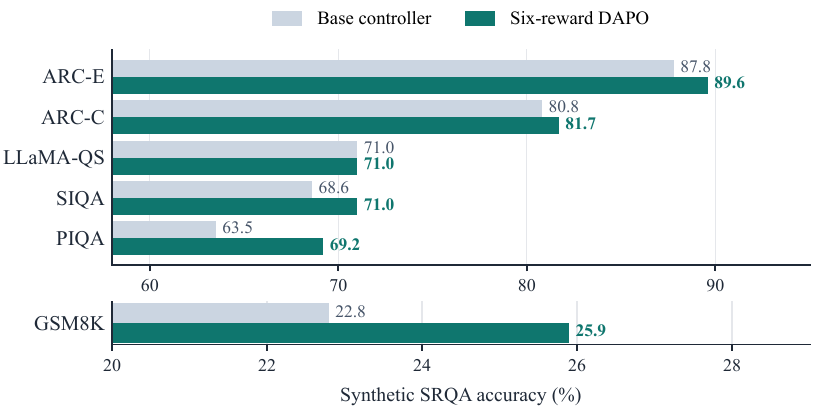}
\caption{Per-task synthetic SRQA accuracy for the base controller and the
six-reward DAPO controller. GSM8K is shown in a separate panel because its
accuracy lies near 25\%.}
\label{fig:per_task_paired}
\end{figure}

Within the Qwen streaming-controller family, the SFT controller provides the
initialization for DAPO policy optimization, while the DAPO rows compare
reward-stack variants trained with four, five, and six reward terms. The SFT row
should be read as protocol learning under token-level supervision, not
direct task-reward optimization. The DAPO stage supplies the task-level reward
that moves the controller back toward task-aligned behavior. The six-reward DAPO controller is the
strongest accuracy row in this block and is second only to the complete-audio
Qwen row in the full table. It improves the synthetic average from 67.6\% for
the base controller and 66.1\% for the SFT controller to 70.3\%, with gains on ARC-C, ARC-E,
SIQA, PIQA, and GSM8K. It also reduces mean post-endpoint
final-think length from 10.44 to 8.99 tokens relative to the base controller.
Figure~\ref{fig:per_task_paired} visualizes the per-task accuracy comparison
between the two controllers, and Appendix~\ref{app:results} reports the
ablation summary across reward stacks.

\subsection{Real human-recorded audio}

Table~\ref{tab:real_audio} evaluates the same controller family on the
human-recorded real-audio benchmark. The SFT controller improves base accuracy
by 4.8 points while staying close to the base final-think length. The DAPO
variants occupy different points along the accuracy--residual-latency
trade-off: the five-reward row has the strongest real-audio accuracy among DAPO
variants, while the six-reward row has the shortest final-think length and is
the only learned controller below the base.

\begin{table}[t]
\caption{Results on Real Audio Bench. Accuracy is computed over 186
human-recorded recordings collected for this work. Final $\downarrow$ reports
mean post-endpoint final-think length, and $\Delta$ columns compare each
controller with the base controller.}
\label{tab:real_audio}
\centering
\small
\renewcommand{\arraystretch}{1.04}
\setlength{\tabcolsep}{4.6pt}
\begin{tabular}{@{}llrrrr@{}}
\toprule
Method & Training signal & Acc. $\uparrow$ & $\Delta$Acc. & Final $\downarrow$ & $\Delta$Final \\
\midrule
Base controller & -- & 64.0 & -- & 6.52 & -- \\
SFT controller & SFT & \textbf{68.8} & \textbf{+4.8} & \underline{6.64} & \underline{+0.12} \\
DAPO controller (4 rewards) & $R_a+R_f+R_s+R_u$ & 65.6 & +1.6 & 7.74 & +1.22 \\
DAPO controller (5 rewards) & $+R_t$ & \underline{67.7} & \underline{+3.7} & 7.39 & +0.87 \\
DAPO controller (6 rewards) & $+R_t+R_c$ & 65.1 & +1.1 & \textbf{6.33} & \textbf{-0.19} \\
\bottomrule
\end{tabular}
\end{table}

Real Audio Bench clarifies the scope of the claim: controller training transfers
beyond TTS-rendered speech, and the SFT and DAPO rows expose different
operating points along the accuracy--residual-latency trade-off. On this
186-item benchmark, the 95\% bootstrap confidence intervals overlap
substantially across learned controllers
(Appendix~\ref{app:realaudio_ci}). We therefore treat
Real Audio Bench as a compact transfer check, not a fine-grained
controller ranking benchmark. The primary accuracy--residual-latency claim
comes from the larger synthetic SRQA evaluation.

\subsection{Controller behavior}
\label{sec:controller_behavior}

Policy optimization changes the trace-level operating point of the streaming
system. The SFT controller initialization sits at 66.1\% synthetic accuracy
with 9.82 final-think tokens. The six-reward DAPO controller reaches 70.3\%
synthetic accuracy with 8.99 final-think tokens, the strongest accuracy and
the shortest synthetic final-think length among the controller variants in
Table~\ref{tab:synthetic_main}. On Real Audio Bench, the controller family
remains functional beyond TTS-rendered speech, but the best accuracy and
shortest final-think points come from different learned variants
(Table~\ref{tab:real_audio}). These shifts show that wait-think-answer control
learns where to place reasoning in the stream.

\subsection{Reproducibility}
\label{sec:repro}
Implementation details are documented in
Sections~\ref{sec:method}--\ref{sec:experiments} (formulation, training
recipe, evaluation protocol) and
Appendices~\ref{app:train}--\ref{app:results} (hyperparameters, reward
definitions, data-generation contract, and ablations).

\section{Limitations}
\label{sec:limitations}

Real Audio Bench is a small five-speaker, 186-recording transfer benchmark;
scaling to a broader user study across accents, environments, and interaction
styles is the planned follow-up. Latency measurements are within-harness only,
separating residual reasoning length (post-endpoint final-think) from
implementation runtime (RTF audit, Appendix~\ref{app:runtime}). Our
implementation uses full-prefix replay, so the RTF audit measures the paper
implementation rather than an ideal cache-native server that would forward
the audio KV cache and avoid repeated prefill. Cache-native serving, larger
real-audio validation, and a head-to-head implementation of
question-completeness controllers in the same Qwen harness are useful future
work; we do not claim a fully optimized production speech system.

\paragraph{Broader impacts.}
Learning when a language model should think during streaming audio has
implications beyond benchmark performance. Low-latency spoken reasoning
supports accessibility tools, real-time captioning, on-device translation,
and conversational tutoring, where response delay disproportionately harms
users who rely on these aids. The same capability admits misuse: a
controller reasoning over partial speech is a step toward systems that
monitor or pre-empt conversations without user awareness, lowers the cost of
real-time social engineering, and can produce asymmetric advantage against
parties without comparable systems. Our design constrains the misuse
surface: the action space is restricted to wait/think/answer with a hard
format gate, and the reward explicitly penalizes spurious thinking,
aligning the controller with non-interruptive conversational norms.

\section{Conclusion}
\label{sec:conclusion}

We formulate streaming speech reasoning as wait-think-answer control for
LALMs: the model learns when to externalize intermediate reasoning during
the audio stream, shifting the accuracy--residual-latency operating point
without changing the base architecture.

\appendix

\section{Benchmark task and data details}
\label{app:benchmark}

The synthetic spoken SRQA benchmark contains six task families. ARC-Easy and
ARC-Challenge are multiple-choice science questions. PIQA tests physical
commonsense, SocialIQA tests social commonsense, GSM8K tests grade-school math,
and LLaMA-QS tests short-answer factual question answering. The row counts in the reported
benchmark are: 2376 ARC-Easy, 1172 ARC-Challenge, 1838 PIQA, 1954
SocialIQA, 1319 GSM8K, and 300 LLaMA-QS examples, for 8,959 total rows.

All tasks are converted into spoken prompts. Multiple-choice tasks are rewritten
so that the options are naturally spoken. GSM8K questions are kept close to the
original wording where possible, with spoken-friendly normalization of notation.
LLaMA-QS items are evaluated as short-answer factual questions. Each rendered
audio file is evaluated under the same benchmark harness used by the controller
rows.

The real-audio benchmark contains 186 final rows after screening and correcting
an initial 200-recording candidate set. The construction process removed
ambiguous or unanswerable items and corrected answer keys before the final
benchmark pass. Qwen controller lanes, Qwen offline lanes, and the retained
external offline baselines all completed on this same corrected evaluation set.

\section{Training, inference, and compute details}
\label{app:train}

We summarize the main training settings for SFT and DAPO. SFT uses MS-Swift
\cite{swift2024} LoRA training \cite{lora2022}; DAPO
uses a custom streaming controller trainer.

\begin{table}[!htbp]
\caption{Training hyperparameters for supervised fine-tuning and DAPO
controller optimization.}
\label{tab:training_hparams}
\centering
\small
\renewcommand{\arraystretch}{1.05}
\setlength{\tabcolsep}{4.2pt}
\resizebox{\linewidth}{!}{%
\begin{tabular}{@{}lll@{}}
\toprule
Setting & Supervised fine-tuning & DAPO controller optimization \\
\midrule
Base model & Qwen2.5-Omni-7B initialized from base model weights & Qwen2.5-Omni-7B initialized from the SFT LoRA adapter \\
Training framework & MS-Swift 4.1.2 & Custom streaming-controller trainer \\
Adaptation & LoRA, rank 8, alpha 32, dropout 0.05 & LoRA actor, rank 8, alpha 32 \\
Target modules & All linear layers & All linear layers \\
Precision & bfloat16 & bfloat16 \\
Hardware & 4 NVIDIA B200 GPUs & 4--5 NVIDIA B200 GPUs, depending on reward stack \\
Maximum length & 8192 tokens & 8192-token rollout context; 48-token think and answer caps \\
Learning rate & $1\times10^{-5}$ & $4\times10^{-7}$ actor learning rate \\
Batching & Per-device batch 4, gradient accumulation 2; effective batch 32 rows/update & 8 rollouts \\
Schedule & One-epoch SFT with cosine decay and warmup & 1000 steps, 50 warmup steps \\
Optimizer & Fused AdamW, weight decay 0.1, betas 0.9/0.95 & AdamW actor update \\
Frozen modules & Audio encoder and aligner frozen; language model trainable & Same base frozen modules as controller LoRA training \\
Model selection & Periodic validation using loss and token accuracy & DAPO variants selected by validation behavior \\
Policy regularization & -- & KL coefficient 0.01; asymmetric clipping 0.20 / 0.28 \\
Reward weights & -- & $\lambda_a=1.0$, $\lambda_f=1.0$, $\lambda_s=1.0$, $\lambda_u=3.0$, $\lambda_t=1.0$, $\lambda_c=0.45$ \\
\bottomrule
\end{tabular}
}
\end{table}

\subsection{Replay runtime audit}
\label{app:runtime}

Table~\ref{tab:runtime_rtf} reports an implementation-level runtime audit for
the reported full-prefix deployment harness. RTF is total controller
wall-clock time divided by source-audio duration. These measurements include
the repeated prefix replay used in the reported controller implementation, so
they complement the final-think token metric and do not replace it. The SFT
initialization controller is the supervised checkpoint used to initialize the DAPO run
in this audit.

\begin{table}[!htbp]
\caption{Replay-harness RTF for the full-prefix deployment audit. Lower is faster.}
\label{tab:runtime_rtf}
\centering
\scriptsize
\renewcommand{\arraystretch}{1.04}
\setlength{\tabcolsep}{3.0pt}
\resizebox{\linewidth}{!}{%
\begin{tabular}{@{}lrrrrrrrr@{}}
\toprule
Controller lane & LLaMA-QS & Real Audio & ARC-C & ARC-E & SIQA & PIQA & GSM8K & Mean \\
\midrule
Base controller & 1.317 & 1.101 & 1.091 & 1.112 & 1.088 & 1.097 & 1.155 & 1.137 \\
SFT controller & 1.584 & 1.179 & 1.317 & 1.362 & 1.316 & 1.376 & 1.336 & 1.353 \\
DAPO controller & 1.278 & 1.136 & 1.080 & 1.198 & 1.035 & 1.088 & 1.123 & 1.134 \\
\bottomrule
\end{tabular}
}
\end{table}

The DAPO controller in this audit is the reported six-reward deployment
lane. Its replay-harness RTF is below the SFT controller and nearly matches
the base controller on average.
This is why the main text treats final-think length as a residual reasoning
metric and reports RTF separately as an implementation diagnostic.

\subsection{Stateful KV-cache controller prototype}

To check whether the same wait-think-answer information flow can be implemented
cache-natively, we built an experimental stateful controller. Audio chunks are
appended into a single persistent Qwen2.5-Omni
thinker KV cache. Each controller decision is generated from a temporary fork
of that persistent state. \texttt{<wait/>} actions are not committed back into
the persistent text cache; \texttt{<think>...</think>} updates are normalized
and appended back; the final-think and answer turns are generated from the
full cached audio plus all committed thoughts.

The prototype runs at the level of
\texttt{model.forward(..., past\_key\_values=..., use\_cache=True)} with a
manual short-token decode loop, instead of the high-level
\texttt{generate()} calls used by the production benchmark path. Mechanism
checks on a four-second real-audio sample confirm that cache lengths match
attention lengths under both 2.0s and 0.5s chunking, and that wait actions
correctly do not enter the persistent cache. These are wiring-level smoke
runs, not benchmark scores; turning the prototype into a score-producing
batch runner remains future work. We treat the prototype as evidence that
the same wait-think-answer information flow can be implemented in a
cache-native deployment without changing the controller's training-time
semantics.

\subsection{SFT and DAPO training details}

\begin{figure}[!htbp]
\centering
\includegraphics[width=0.82\linewidth]{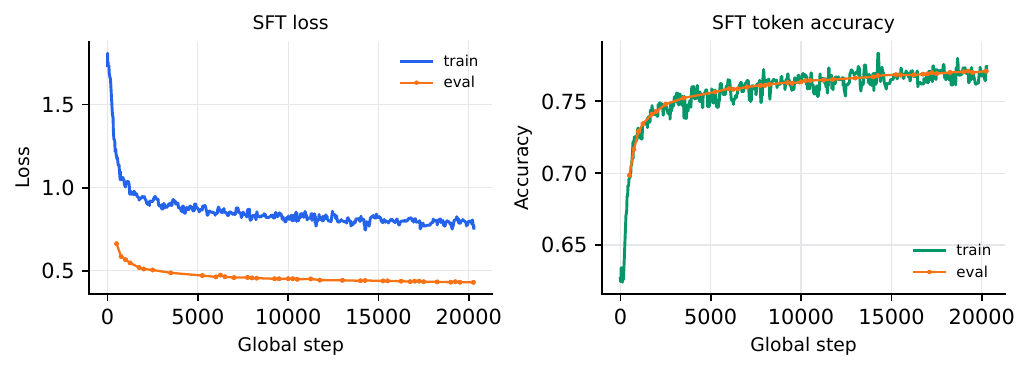}
\caption{SFT training curves from the audio-only cold-start run. Training and
validation metrics improve during supervised fine-tuning, providing the
initialization for policy optimization.}
\label{fig:sft_curves}
\end{figure}

The audio-only cold-start SFT run shows decreasing training loss and increasing token
accuracy during training (Figure~\ref{fig:sft_curves}). DAPO update points are
chosen based on validation behavior, since the streaming
reward is non-monotonic across policy updates. For the controller tick
sweep, 0.5s, 1.0s, and 1.5s grids achieved validation token accuracies of
0.6108, 0.6105, and 0.6102, respectively; we use 0.5s because accuracy was tied
while timing resolution was finer. Qwen2.5-Omni calls are still padded or
extended to the 2.0s minimum audio window required by the model, so the sweep
changes the reconsideration cadence while leaving the minimum audio exposure
fixed.

For DAPO, each prompt is sampled into a group of $G$ rollouts. We compute
group-relative advantages
\[
A_i=\frac{R_i-\frac{1}{G}\sum_{j=1}^{G}R_j}
{\operatorname{std}(R_1,\ldots,R_G)+\epsilon},
\]
and apply token-level clipped updates over the completion-token mask $M$. With
$y_{i,t}$ denoting the $t$-th generated token in rollout $i$, $o_{i,t}$ its
token-level context, and
$r_{i,t}(\theta)=\exp(\log\pi_\theta(y_{i,t}\mid o_{i,t})-
\log\pi_{\mathrm{old}}(y_{i,t}\mid o_{i,t}))$, the objective is
\begin{equation}
\mathcal{L}_{\mathrm{DAPO}} =
-\frac{1}{|M|}
\sum_{(i,t)\in M}
\begin{cases}
\min(r_{i,t}A_i,\operatorname{clip}(r_{i,t},1-\epsilon_l,1+\epsilon_h)A_i),
& A_i\geq 0,\\
\max(r_{i,t}A_i,\operatorname{clip}(r_{i,t},1-\epsilon_l,1+\epsilon_h)A_i),
& A_i<0.
\end{cases}
\end{equation}
Before a DAPO update, sampled groups must contain enough format-valid rollouts,
at least one valid final-think rollout, and at least one valid pre-endpoint
thought. Groups that collapse to all-wait or malformed final-think behavior are
resampled up to a fixed retry budget; persistent failures are skipped and logged.
Table~\ref{tab:reward_details_appendix} gives the operational definitions used
by the reward implementation.

\begin{table}[H]
\caption{Operational definitions of reward terms. Positive signals define the
intended controller behavior, while failure modes describe behavior penalized
during DAPO.}
\label{tab:reward_details_appendix}
\centering
\scriptsize
\renewcommand{\arraystretch}{1.12}
\setlength{\tabcolsep}{3.0pt}
\begin{tabular}{@{}p{0.06\linewidth}p{0.45\linewidth}p{0.39\linewidth}@{}}
\toprule
Term & Positive signal & Failure mode discouraged \\
\midrule
$R_a$ & Correct final answer after deterministic task scoring; local judge scoring is used only as a semantic-equivalence fallback for open short-answer cases. & Fluent but incorrect final answers. \\
$R_f$ & Valid wait-think-answer format: wait/think while listening, one final think, and answer only after the utterance ends. & Early answers, malformed tags, missing final-think turns, or invalid action order. \\
$R_s$ & Short post-endpoint final-think state once the audio stream is complete. & Repeating reasoning after the endpoint that should have been resolved during listening. \\
$R_u$ & Intermediate thoughts placed near answer-relevant evidence updates in the audio. & Missing an important state change or interrupting the stream with irrelevant thoughts. \\
$R_t$ & Local judge preference for short, concrete, answer-supporting semantic states. & Generic, verbose, or meta-level thoughts that do not update the answer state. \\
$R_c$ & Local judge preference for thought chains that support the final answer when the answer is correct. & Internally inconsistent chains or unsupported jumps to the final answer. \\
\bottomrule
\end{tabular}
\end{table}

\paragraph{Reward implementation details.}
The six terms above are the conceptual reward categories; the implementation
adds only local guardrails inside these categories. For $R_a$, deterministic
scoring is used whenever possible: answers are normalized, multiple-choice
labels are mapped to option text when possible, and numeric answers are compared
after lightweight numeric normalization. Local judge scoring is used only as a
semantic-equivalence fallback for open short-answer cases where exact matching
is insufficient. The same term uses a difficulty-aware effort calibration, so a
correct answer is not rewarded equally when the controller uses an implausibly
empty or unnecessarily verbose reasoning state.
For $R_s$, the measured quantity is the post-endpoint final-think length. The
runtime token count is used when available; otherwise the reward path falls back
to a character-based token estimate. The reported configuration gives the
final-think state a six-token free budget and then applies a linear penalty with
coefficient $0.30$, capped at $3.0$. Correct and well-formed trajectories can
receive a small bonus when the final-think state is a compact answer cue of
three to six tokens; this bonus is disabled when the answer shape is invalid.
The answer-shape guardrail penalizes question-form answers, label-only answers
when a semantic answer is expected, and yes/no type mismatches.

For $R_u$, pre-endpoint thoughts are matched to answer-relevant evidence-update
ticks derived from the controller metadata. Matching uses a tolerance of two
0.5s ticks and includes sparsity pressure, so the controller is penalized both
for missing important updates and for inserting thoughts at nearly every tick.
The judge-assisted terms use a local Qwen3.6-35B-A3B endpoint
\cite{qwen36_35b_a3b2026}. $R_t$ scores emitted thoughts as local semantic
state updates, and $R_c$ scores whether the visible thought chain supports the
final answer. Judge outputs are mapped to coarse scores in $\{0,0.5,1\}$ and are
used only as reward signals; the full judge prompts are part of the released
training code rather than the main paper text.

\paragraph{Reward weight selection.}
The reward weights in Table~\ref{tab:training_hparams} were selected through
validation-set sweeps during method development. Among the soft-coefficient
terms, the update-timing weight $\lambda_u$ and the latency weight $\lambda_s$
had the largest qualitative effect on the accuracy--residual-latency operating point
because they directly shape when and how often the controller emits visible
reasoning state. The correctness, protocol, thought-quality, and
chain-consistency weights ($\lambda_a, \lambda_f, \lambda_t, \lambda_c$) were
comparatively less sensitive in the explored range, with $\lambda_f$ acting
mainly as a protocol gate once it exceeded the threshold needed for valid
action format and $\lambda_t, \lambda_c$ providing soft preferences over
judge-scored signals. The reported configuration corresponds to the
validation-best operating point under the synthetic SRQA setup.

\section{Prompts, alignment, and data construction notes}
\label{app:prompts}

The canonical synthetic record contains the fields \texttt{question\_text},
\texttt{tts\_text}, \texttt{tts\_instruct}, \texttt{transcript\_text},
\texttt{anchor\_words}, \texttt{logical\_actions}, and
\texttt{final\_answer}. Here \texttt{question\_text} is the canonical task
question, \texttt{tts\_text} is the spoken surface form passed to the TTS
renderer, \texttt{tts\_instruct} is the delivery/style control prompt,
\texttt{transcript\_text} is the normalized transcript aligned to the waveform,
\texttt{anchor\_words} are lexical anchors used to localize answer-relevant state
changes, \texttt{logical\_actions} is the semantic wait-think-answer trace before
serialization, and \texttt{final\_answer} is the gold answer string. The teacher
prompt requires spoken-friendly wording,
avoids unnatural written math notation where possible, and asks for thoughts
that preserve short semantic state instead of generic summaries, tone
descriptions, or pause descriptions. Acoustic style is controlled through
\texttt{tts\_instruct}; it should influence the rendered speech, not become the
content of the reasoning state unless it changes the answer.

Full spoken inputs are rendered once. We do not synthesize each controller segment
independently, because segment-wise synthesis would inject artificial prosodic
breaks exactly where the model is supposed to learn natural timing. Forced
alignment maps transcript words to audio timestamps, and action anchors are
snapped upward to the 0.5s controller grid. The supervised export and the DAPO
export are both generated from this aligned representation.

\subsection{Synthetic data generation prompt}

The project-specific synthetic data prompt is implemented in the data-generation
code and mirrored in the reviewer-facing prompt document. The core GPT-4o
\cite{gpt4o2024} contract is:

{\footnotesize
\begin{verbatim}
System:
You create high-quality English training records for a streaming speech
reasoning model. Return one JSON object only.

The spoken script is always user speech only. Do not solve the task or speak as
the assistant inside tts_text or transcript_text. Do not use inline TTS markup
such as <break>. Use punctuation and a separate tts_instruct field to induce
natural speech style when needed.

transcript_text must be clean English alignment text. logical_actions must use
only <wait/>, <think>...</think>, and <answer>...</answer>. Some anchors may map
to <wait/> because a physical pause or filler does not necessarily require
thinking.

Every <think> should be short, incremental, grounded in the current evidence,
and useful for the answer state. Prefer concrete facts, constraints,
eliminations, quantities, corrections, answer-type constraints, or current
candidate state. Avoid generic meta commentary and avoid long explanations.

Every sample must be unique. Treat the scenario seed only as a starting
situation and vary names, numbers, locations, items, and wording. Do not use
LaTeX, bullet lists, stage directions, or duplex terminology.

User template:
Create one English training example.
sample_id: {sample_id}
task_family: {task_family}
topic: {topic}
verifiability: {verifiability}
task_type: {task_type}
difficulty: {difficulty}
variant_index: {variant_index}
scenario_seed: {scenario_seed}
Target about {pause_count} meaningful anchor boundaries.
{family_guidance}
{difficulty_guidance}

Return JSON fields:
task_type, topic, verifiable, difficulty, question_text, tts_text,
tts_instruct, transcript_text, anchor_words, logical_actions, final_answer,
difficulty_metadata.

Key requirements:
- tts_text and transcript_text contain only user speech.
- anchor_words are copied from transcript_text.
- logical_actions has one key per anchor plus anchor_AUDIO_END.
- non-final anchors use <wait/> or <think>...</think>, never <answer>.
- anchor_AUDIO_END contains one final <think>...</think> followed by
  one <answer>...</answer>.
- the answer text exactly matches final_answer.
\end{verbatim}
}

\subsection{Spoken SRQA prompt provenance}

For the six synthetic SRQA benchmark families, we follow the spoken-rewrite
prompt contract from Shih et al.~\cite{shih2026speech}. The implementation uses
their task-specific prompts for ARC, PIQA, SocialIQA, and GSM8K style
conversion: preserve all information needed for the answer, verbalize choices
inside the spoken question, and return a JSON object containing the converted
spoken question and, where applicable, the converted answer. We cite their
prompt without reproducing it as a new project-specific prompt.

\subsection{Real Audio Bench generation prompt}

Real Audio Bench uses a separate GPT-4o candidate-generation prompt because its
goal is real human delivery, not TTS rendering. The source generation spec fixes
200 planned items, five speakers, and 40 recordings per speaker, then filters
the recorded set to 186 final rows. The universal prompt is:

{\footnotesize
\begin{verbatim}
System:
You are writing candidate items for a real-human benchmark for streaming spoken
reasoning.

This benchmark is NOT ordinary QA. It is for a streaming controller that hears
audio over time and must wait until a late decisive cue arrives.

Hard constraints:
- Write natural spoken English only.
- The utterance must sound like something a person would actually say aloud to
  an assistant.
- Every item must be verifiable from the utterance itself.
- Each item must have exactly one preferred gold answer.
- The spoken utterance must contain an early plausible answer that later
  becomes wrong, incomplete, or premature.
- A late cue must overwrite or finalize the answer.
- Avoid textbook math, school trivia, science, history, and canned benchmark
  style.
- Keep spoken duration within 8 to 18 seconds.
- Keep the answer short and cleanly scoreable.
- Across a batch, diversify the reasoning mechanism.

User template:
Generate {count} candidate items.

Coverage requirements:
- use the requested categories, scenario families, difficulties, and reasoning
  mechanisms as evenly as possible
- all items must be verifiable
- prioritize distinctness from the existing synthetic corpus
- every item must pass a uniqueness test: after reading the full utterance,
  there should be exactly one reasonable gold answer
- when using time windows, thresholds, eligibility, or scheduling conflicts,
  state the deciding rule explicitly

Allowed categories:
mid_way_reversal, adversarial_pause, incremental_accumulation,
noisy_environment, fast_speech_hesitation, real_world_situational

Allowed scenario families:
day_of_travel_logistics, home_landlord_appliance, workplace_school_admin_it,
food_shopping_returns, health_pharmacy_finance_lite

Allowed reasoning mechanisms:
overwrite_final_slot, cumulative_total, tiered_discount_total,
fee_or_threshold_decision, exclusion_choice, bounded_window_selection,
quantity_update, eligibility_decision, refund_or_credit_total,
schedule_window_resolution

Return one JSON object with an items array. Each item has:
item_id, category, difficulty, scenario_family, reasoning_mechanism,
verifiability, question_spoken, gold_answer, answer_check_type,
why_unique_answer, ambiguity_check, why_early_answer_would_be_wrong,
critical_audio_cue, speaker_notes, estimated_duration_sec, acoustic_target.
\end{verbatim}
}

After generation, an LLM verifier and human audit screened the candidate pool
for a unique answer, spoken naturalness, reasoning strength, and distinctness.
The final manifest keeps 186 recorded items, drops 14 ambiguous or unsuitable
items, and applies 23 audited answer-key corrections.

\section{Additional results and ablations}
\label{app:results}

Table~\ref{tab:controller_variants} summarizes the Qwen-family controller
variants used in the main results table, separating the SFT controller from
the four-, five-, and six-reward DAPO controllers on synthetic SRQA accuracy
and mean final-think length.

\begin{table}[!htbp]
\caption{Ablation summary for Qwen controller variants on synthetic SRQA. The
table reports the base controller, the SFT controller, and DAPO controllers trained with
four, five, or six reward terms.}
\label{tab:controller_variants}
\centering
\small
\renewcommand{\arraystretch}{0.98}
\setlength{\tabcolsep}{4.0pt}
\begin{tabular}{@{}llrr@{}}
\toprule
Variant & Reward stack & Syn. acc. $\uparrow$ & Syn. final $\downarrow$ \\
\midrule
Base controller & -- & 67.6 & 10.44 \\
SFT controller & SFT & 66.1 & \underline{9.82} \\
DAPO controller (4 rewards) & $R_a+R_f+R_s+R_u$ & 68.5 & 10.87 \\
DAPO controller (5 rewards) & $R_a+R_f+R_s+R_u+R_t$ & \underline{69.2} & 10.94 \\
DAPO controller (6 rewards) & $R_a+R_f+R_s+R_u+R_t+R_c$ & \textbf{70.3} & \textbf{8.99} \\
\bottomrule
\end{tabular}
\end{table}

\subsection{Real Audio Bench bootstrap confidence intervals}
\label{app:realaudio_ci}

Real Audio Bench contains 186 human-recorded items
(Section~\ref{sec:experiments}). To quantify the uncertainty introduced by
this small evaluation size, Table~\ref{tab:realaudio_ci} reports per-lane
accuracy with 95\% bootstrap confidence intervals over 10{,}000 resamples of
the 186 items. The lanes match the controller variants reported in
Table~\ref{tab:real_audio}.

\begin{table}[H]
\caption{Real Audio Bench accuracy with 95\% bootstrap confidence intervals
(10{,}000 resamples). The lanes match the controller variants reported in
Table~\ref{tab:real_audio}.}
\label{tab:realaudio_ci}
\centering
\small
\renewcommand{\arraystretch}{1.06}
\setlength{\tabcolsep}{5.5pt}
\begin{tabular}{@{}lrrl@{}}
\toprule
Lane & Correct/Total & Acc. & 95\% bootstrap CI \\
\midrule
Base controller & 119/186 & 0.640 & [0.570, 0.710] \\
SFT controller & 128/186 & 0.688 & [0.618, 0.753] \\
DAPO controller (4 rewards) & 122/186 & 0.656 & [0.586, 0.726] \\
DAPO controller (5 rewards) & 126/186 & 0.677 & [0.608, 0.742] \\
DAPO controller (6 rewards) & 121/186 & 0.651 & [0.581, 0.720] \\
\bottomrule
\end{tabular}
\end{table}

The 95\% confidence intervals overlap substantially across all controller
lanes, including between the base controller and the post-trained variants.
We therefore treat Real Audio Bench primarily as a transfer check, not
a fine-grained controller ranking benchmark. Per-controller synthetic numbers
are summarized in Table~\ref{tab:controller_variants}.

\section{Asset licenses}
\label{app:licenses}

Table~\ref{tab:licenses} summarizes the public licenses of the upstream
assets used in this work. All assets are accessed through their official
release channels and used within their published terms. License or terms
information was checked against each asset's official model card, dataset
card, repository, \texttt{LICENSE} file, or source documentation where
available, at the time of submission.

\begin{table}[H]
\caption{Public licenses or terms of use for upstream assets used in this work.}
\label{tab:licenses}
\centering
\small
\renewcommand{\arraystretch}{1.05}
\setlength{\tabcolsep}{5.5pt}
\begin{tabular}{@{}lll@{}}
\toprule
Asset & Type & License \\
\midrule
Qwen2.5-Omni-7B \cite{qwen2025omni} & Model & Apache 2.0 \\
Qwen3-TTS \cite{qwen3tts2026} & Model & Apache 2.0 \\
Qwen3.6-35B-A3B \cite{qwen36_35b_a3b2026} & Model & Apache 2.0 \\
GLM-4-Voice-9B \cite{glm4voice2024} & Model & GLM-4 Model License Agreement; code under Apache 2.0 \\
Audio Flamingo 3 \cite{af32025} & Model & NVIDIA OneWay Noncommercial License \\
GPT-4o \cite{gpt4o2024} & API & OpenAI API terms of service \\
MS-Swift \cite{swift2024} & Software & Apache 2.0 \\
ARC \cite{arc2018} & Dataset & CC-BY-SA 4.0 \\
PIQA \cite{piqa2020} & Dataset & Academic Free License 3.0 \\
SocialIQA \cite{socialiqa2019} & Dataset & CC-BY 4.0 \\
GSM8K \cite{gsm8k2021} & Dataset & MIT \\
LLaMA-Questions/Spectron \cite{spectron2023} & Dataset & Source paper / archived GitHub terms; license not explicitly specified \\
\bottomrule
\end{tabular}
\end{table}

\end{document}